\documentclass[sn-mathphys,Numbered]{sn-jnl}

\usepackage{graphicx}%
\usepackage{multirow}%
\usepackage{amsmath,amssymb,amsfonts}%
\usepackage{amsthm}%
\usepackage{mathrsfs}%
\usepackage[title]{appendix}%
\usepackage{xcolor}%
\usepackage{textcomp}%
\usepackage{manyfoot}%
\usepackage{booktabs}%
\usepackage{algorithm}%
\usepackage{algorithmicx}%
\usepackage{algpseudocode}%
\usepackage{listings}%

\usepackage{times}
\usepackage{latexsym}
\usepackage{times}
\usepackage{latexsym}
\usepackage{amsfonts}
\usepackage{amsmath}
\usepackage{amssymb}
\usepackage{multicol}
\usepackage{multirow}
\usepackage{xspace}
\usepackage{booktabs}
\usepackage{bbding}
\usepackage{array}
\usepackage{threeparttable}
\usepackage{tcolorbox}
\usepackage{enumitem}
\usepackage{subfigure}
\usepackage{xcolor,colortbl}

\newcommand{\xhdr}[1]{\vspace{0.3em}\noindent{{\bf #1.}}}

\theoremstyle{thmstyleone}%
\theoremstyle{thmstyletwo}%

\theoremstyle{thmstylethree}%

\begin{document}

\title[Knowledge-Enhanced Relation Extraction Dataset]{Knowledge-Enhanced Relation Extraction Dataset}


\author[1]{\fnm{Yucong} \sur{Lin}}\email{linyucong@bit.edu.cn}\equalcont{These authors contributed equally to this work.}
\author[2]{\fnm{Hongming} \sur{Xiao}}\email{xiaohongmin@bit.edu.cn}\equalcont{These authors contributed equally to this work.}
\author[2]{\fnm{Jiani} \sur{Liu}}\email{jiani\_liu@bit.edu.cn}
\author[2]{\fnm{Zichao} \sur{Lin}}\email{zc\_lin@bit.edu.cn}
\author[3]{\fnm{Keming} \sur{Lu}}\email{keminglu@usc.edu}
\author*[4]{\fnm{Feifei} \sur{Wang}}\email{feifei.wang@ruc.edu.cn}
\author*[5]{\fnm{Wei} \sur{Wei}}\email{weiwei@phbs.pku.edu.cn}

\affil[1]{\orgdiv{School of Medical Technology}, \orgname{Beijing Institute of Technology}, \orgaddress{\street{Zhongguancun South Street No.5}, \city{Beijing}, \postcode{100081}, \country{China}}}

\affil[2]{\orgdiv{School of Computer Science and Technology}, \orgname{Beijing Institute of Technology}, \orgaddress{\street{No.5, Zhongguancun South Street}, \city{Beijing}, \postcode{100081}, \country{China}}}

\affil[3]{\orgdiv{Viterbi School of Engineering, University of Southern California}, \orgname{University of Southern California}, \orgaddress{\street{3939 S Figueroa St}, \city{Los Angeles}, \state{CA}, \postcode{90037}, \country{USA}}}

\affil[4]{\orgdiv{Center for Applied Statistics and School of Statistics}, \orgname{Renmin University of China}, \orgaddress{\street{No. 59, Zhongguancun Street}, \city{Beijing}, \postcode{100872}, \country{China}}}

\affil[5]{\orgdiv{HSBC Business School}, \orgname{Peking University}, \orgaddress{\street{Xueyuan Street}, \city{Shenzhen}, \postcode{518055}, \state{Guangdong} \country{China}}}


\abstract{
Recently, knowledge-enhanced methods leveraging auxiliary knowledge graphs have emerged in relation extraction, surpassing traditional text-based approaches. 
However, to our best knowledge, there is currently no public dataset available that encompasses both evidence sentences and knowledge graphs for knowledge-enhanced relation extraction. 
To address this gap, we introduce the Knowledge-Enhanced Relation Extraction Dataset (KERED). 
KERED annotates each sentence with a relational fact, and it provides knowledge context for entities through entity linking. 
Using our curated dataset, We compared contemporary relation extraction methods under two prevalent task settings: sentence-level and bag-level. 
The experimental result shows the knowledge graphs provided by KERED can support knowledge-enhanced relation extraction methods.
We believe that KERED offers high-quality relation extraction datasets with corresponding knowledge graphs for evaluating the performance of knowledge-enhanced relation extraction methods. 
Our dataset is available at: \url{https://figshare.com/projects/KERED/134459}
}


\maketitle

\section{Introduction}\label{introduction}
Relation extraction (RE) focuses on extracting relationships between entities from natural language sentences \cite{DBLP:journals/corr/abs-1712-05191}. 
RE enhances various downstream tasks in natural language processing, including question answering \cite{xu-etal-2016-question, chen-etal-2019-uhop}, knowledge graph construction \cite{luan-etal-2018-multi, bosselut-etal-2019-comet}, and reading comprehension \cite{qin-etal-2021-erica, DBLP:journals/corr/abs-2006-11880}. 
Knowledge graphs (KGs) store relational facts as triples including subject entities, object entities, and the relations between them \cite{9416312}. 
For instance, the relational fact \emph{(James Joyce, country of citizenship, Ireland)} indicates that James Joyce was a citizen of Ireland. 
As a kind of structured representation of facts, KGs find extensive applications, such as social network analysis \cite{7358050, DBLP:journals/corr/abs-1807-00504} and recommender systems \cite{10.1145/3178876.3186175, Zou_2020}.

\begin{figure}[ht]
    \centering
    \includegraphics[width=\linewidth]{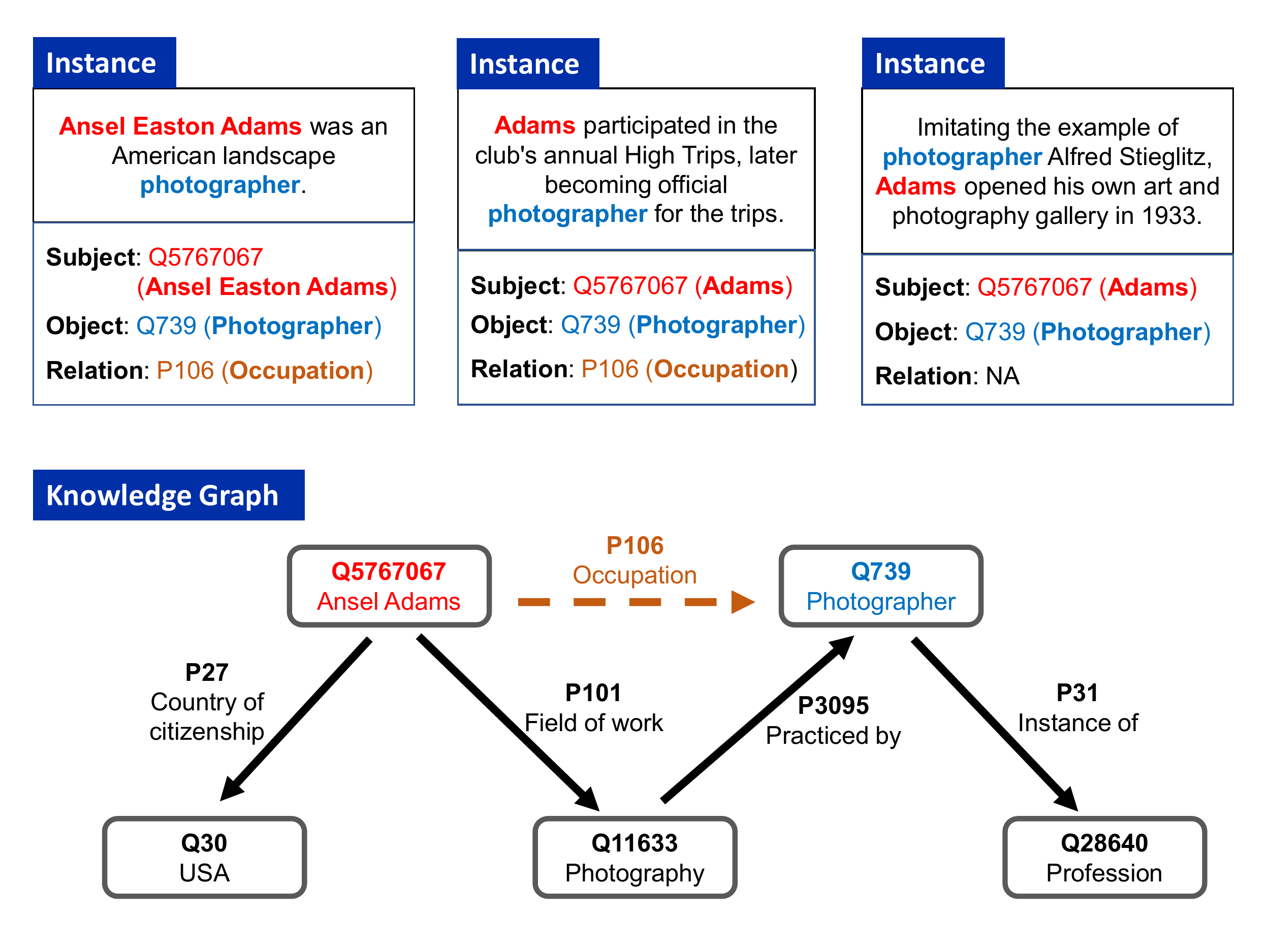}

    \caption{An illustration of instances in a DSRE dataset and the auxiliary KG for the dataset. An instance consists of a sentence together with a relational fact expressed by the sentence. A batch of instances constitutes an RE dataset. A KG contains some or all entities in an RE dataset and relations between them. In this example, we link entities in the three instances to their counterparts in the auxiliary KG. Then, a knowledge-enhanced RE method can classify the relation between \emph{Ansel Adams} and \emph{Photographer} as \emph{Occupation} by the virtue of the information provided by the sentences and the KG.}
    \label{fig:instance}
\end{figure}

Recently, KGs are widely used as auxiliary information to enhance RE methods \cite{zhang-etal-2019-long,liang2021distantly,nadgeri-etal-2021-kgpool}. 
And, the availability of resources such as distantly supervised relation extraction (DSRE) datasets \cite{gao-etal-2021-manual} and extensive knowledge databases \cite{wikidata, freebase} has facilitated the study of knowledge-enhanced RE. 
However, no public RE dataset exists that aligns sentences with the corresponding knowledge context for training and evaluating knowledge-enhanced RE methods. 
Previous researchers in this field tend to construct their own auxiliary KGs, create datasets from scratch, and retest prior benchmarks for fair comparisons \cite{liang2021distantly, recon, nadgeri-etal-2021-kgpool}. 
The lack of public benchmarks makes it challenging to report reproducible results or compare the performance of existing methods.

To address these issues, we adapt three widely-used RE datasets for knowledge-enhanced RE tasks to curate the \textbf{K}nowledge-\textbf{E}nhanced \textbf{R}elation \textbf{E}xtraction \textbf{D}ataset (KERED).
KERED improves the data quality of previous datasets. Also, with information from external knowledge bases, KERED constructs auxiliary KGs for entities in the dataset. 
We believe KERED will foster the development of knowledge-enhanced RE in the future.

We commenced our work by examining the original DSRE datasets based on Wikidata \cite{wikidata} or Freebase \cite{freebase}. 
Because DSRE generates large-scale data by aligning relational facts in knowledge bases with evidence sentence \cite{gao-etal-2021-manual}, mentions in the corpus naturally match their entities in their corresponding knowledge bases for distant supervision, which streamlines the process of locating entities via their identifiers and access the knowledge context of entities in the knowledge bases. 
Therefore, it is feasible to construct a KG for each DSRE dataset, as illustrated in Figure \ref{fig:instance}. 
Specifically, We collected three DSRE datasets for KERED: NYT10m \cite{gao-etal-2021-manual}, Wiki20m \cite{gao-etal-2021-manual}, and Wiki80 \cite{han-etal-2019-opennre}.
Then, we refined them by enhancing data quality and constructing a KG for each. 
Also, we conducted comprehensive RE experiments on KERED to evaluate the performance of existing RE methods.
The experimental result shows that information from auxiliary KGs has positive effects on RE methods. 
In summary, KERED offers the first standardized datasets for knowledge-enhanced RE tasks.

Our study's contributions are twofold. 
Firstly, we develop KERED, comprising three challenging RE datasets with auxiliary KGs, with the potential to advance knowledge-enhanced RE research. We make our datasets publicly available on Figshare\footnote{\url{https://figshare.com/projects/KERED/134459}}; please refer to Appendix A for KERED access. 
Secondly, we establish metrics for knowledge-enhanced RE methods on KERED and assess state-of-the-art RE methods using our datasets. 
Our experiments indicate that knowledge-enhanced RE methods can surpass traditional approaches. 

The remainder of the paper is organized as follows: Section~\ref{related work} reviews widely-used DSRE datasets and knowledge-enhanced RE methods. Section~\ref{data collection} details the construction of KERED. Section~\ref{data analysis} presents KERED's descriptive analysis. Section~\ref{benchmarks} evaluates RE models on KERED. Section~\ref{discussion} discusses the experimental results. Finally, Section~\ref{conclusion} concludes the paper.

\section{Related Works}\label{related work}
\xhdr{DSRE datasets} The majority of existing DSRE datasets are constructed by identifying entities mentioned in evidence sentences and linking them to public knowledge bases such as Wikidata \cite{wikidata} and Freebase \cite{freebase}. 
NYT10 \cite{nyt10} is a large-scale dataset automatically constructed using DSRE. 
However, \citet{han-etal-2019-opennre} highlighted noisy labeling issues in NYT10 and other existing datasets. 
To mitigate this problem, \citet{gao-etal-2021-manual} introduced two DSRE datasets with manually annotated test sets, significantly enhancing the data quality of previous NYT10 \cite{nyt10} and Wiki20 \cite{wiki20}. 
Although noisy labeling problems were solved in test sets, data quality issues persist in NYT10m and Wiki20m. 
Hence, we further denoised these datasets and enriched them with external KGs through entity linking. 
Our refined datasets provide the community with benchmarks for evaluating knowledge-enhanced RE methods.

\xhdr{Knowledge-enhanced Relation Extraction} An increasing number of RE methods incorporate auxiliary information, such as attributes and embeddings of entities, into their models \cite{vashishth-etal-2018-reside}, where KG information plays a crucial role by revealing associations between entities \cite{christopoulou-etal-2021-distantly}. 
CGRE \cite{liang2021distantly} derives constraint graphs from KGs to model intrinsic connections between relations. 
The model generates representations for entities and relations by encoding the graph into vectors and extracting node features. 
\citet{xu-barbosa-2019-connecting} proposed an RE framework HRERE which jointly learns language representations and knowledge graph embeddings. 
Moreover, KGPool \cite{nadgeri-etal-2021-kgpool} employs a graph pooling algorithm that dynamically selects KG context to enhance model performance. Their method considers only names, categories, aliases, and descriptions of entities as their side information from KG. 
REMAP \cite{lin2022multimodal}, a multimodal method for DSRE, combines knowledge graph embeddings with deep language models to classify relations between entities. 
However, comparing the performance of all these methods is impossible due to the absence of benchmarks. 
Thus, we revisit existing knowledge-enhanced RE methods and evaluate them on our datasets to facilitate an objective comparison.

\section{Construction of KERED}\label{data collection}
We first define the problem and present an overview of previous DSRE datasets in Sections~\ref{sec:definition} and \ref{sec:previous}. 
Subsequently, we detail the process of entity linking in Sections~\ref{sec:linking}, and then describe the process of dataset refinement in \ref{sec:refinement}.

\subsection{Problem Definition}
\label{sec:definition}
An RE dataset consists of a collection of instances and a set of candidate relations.
Wherein, each instance contains an evidence sentence and an annotated relational fact (an entity pair and a relation between them), as illustrated in Figure~\ref{fig:instance}.
All relations appearing in the dataset are restricted to the set of candidate relations.
On RE datasets, traditional RE methods predict the relation between two entities based on the sentence for each instance.
In addition to that, knowledge-enhanced RE methods typically require an auxiliary KG as supplementary information to improve RE performance.
They categorize relations between entity pairs in evidence sentences with the aid of knowledge base information \cite{vashishth-etal-2018-reside, xu-barbosa-2019-connecting}.
To establish datasets for knowledge-enhanced RE tasks, we enhance the data quality of existing RE datasets and construct KGs for them to obtain KG-enhanced RE datasets.

Our experiments evaluate RE methods at two levels.
Sentence-level RE considers only one instance as input at a time and predicts the relation of entities expressed by the sentence.
In contrast, bag-level RE takes a bag as input at a time. 
The bag contains multiple instances with the same entity pairs in the dataset, such as the three instances depicted in Figure~\ref{fig:instance}.
When categorizing the relation between two entities, bag-level RE methods take into account all instances in the bag.

The main notations used are as follows: $\mathcal{E}$ represents the entity set containing all entities in an RE dataset.
$\mathcal{R}$ represents the relation set containing all relation labels in an RE dataset.
N/A means not applicable, which is a label type that may appear in $\mathcal{R}$.
$r^i=(x^i, e_1^i, e_2^i)$ is the $i$-th instance in an RE dataset, where $e^i_1$ and $e^i_2$ delimit entity mentions in token sequence $x^i$.
$(h^j, t^j, r^j)$ is the $j$-th relation fact in a KG, where $h^j$ and $t^j$ are subject entity and object entity with relation $r^j$, respectively.

\subsection{Previous Relation Extraction Datasets}
\label{sec:previous}
We utilized three frequently-used relation extraction datasets: NYT10m \cite{gao-etal-2021-manual}, Wiki80 \cite{han-etal-2019-opennre}, and Wiki20m \cite{gao-etal-2021-manual}.
Wiki80 is a sentence-level dataset, while Wiki20m and NYT10m are bag-level datasets.

\begin{itemize}[leftmargin=1em]
\setlength\itemsep{0em}
\item \textbf{NYT10m} \cite{gao-etal-2021-manual} is a bag-level dataset derived from NYT10 \cite{riedel2010modeling} by cleaning the dataset and separating the validation set from the training set.
NYT10m also provides a manually-annotated test set based on the original test set.
N/A instances in NYT10m indicate that there is no relation between entities.

\item \textbf{Wiki80} \cite{han-etal-2019-opennre} is a sentence-level dataset based on the few-shot dataset FewRel \cite{han-etal-2018-fewrel}.
This human-labeled dataset contains 56,000 relation facts with 80 types of relations.
N/A relation facts do not exist in Wiki80.

\item \textbf{Wiki20m}~\cite{gao-etal-2021-manual} is derived from Wiki20 by reorganizing its relation facts and redividing its training, validation, and test sets.
This bag-level RE dataset shares the same relation ontology with Wiki80, except that Wiki20m contains an N/A relation expressing unknown relations.
\end{itemize}

\subsection{Quality Assessment of Data Source}
\label{sec:quality}
In this section, we discuss the data quality of source datasets in KERED, namely NYT10m, Wiki80, and Wiki20m. Subsequently, we address the data quality of the knowledge bases we employed, specifically Freebase and Wikidata.

Wiki80 is derived from a manually-checked RE dataset FewRel~\cite{han-etal-2018-fewrel}. Each instance in FewRel was reviewed by at least two well-educated annotators to filter incorrectly labeled instances generated by DSRE. Consequently, the data quality of Wiki80 is assured. NTY10m \cite{gao-etal-2021-manual} annotated all non-NA instances in the test set, and the train/validation sets were generated directly by DSRE. Additionally, the authors reorganized the instances in Wiki20 to construct Wiki20m, using Wiki80 as the test set. In comparison with Wiki80, both NYT10m and Wiki20m are large-scale datasets (approximately ten times larger than Wiki80), making it infeasible to manually check them all. Thus, DSRE-generated train/validation sets are acceptable for them. Moreover, as discussed in Section~\ref{sec:refinement}, we further enhanced the data quality of NYT10m, Wiki80, and Wiki20m. Therefore, all RE datasets in KERED are of high quality and suitable for evaluating the performance of RE methods.

As we construct KGs for KERED using external knowledge bases, it is essential to verify their credibility, specifically Freebase and Wikidata. \cite{farber2018linked} analyzes five widely-used knowledge bases, providing a comprehensive framework to evaluate their characteristics and quality. According to the authors, RDF documents and literals in both Freebase and Wikidata are syntactically verified by editors. Furthermore, both knowledge bases provide the provenance of each relational fact to ensure the statement's validity. The general accuracy of relational facts in Freebase and Wikidata exceeds 99.3\% \cite{farber2018linked}. In summary, the accuracy and trustworthiness of the knowledge sources of KGs in KERED are robust and guaranteed.

\subsection{Entity Linking}
\label{sec:linking}
Entity linking involves locating entities in a knowledge base to extract the knowledge context for each entity, enabling the construction of a KG for a given dataset.
For KERED, we applied different entity linking methods to NYT10m~\cite{gao-etal-2021-manual}, Wiki80~\cite{han-etal-2019-opennre}, and Wiki20m~\cite{gao-etal-2021-manual}.

\xhdr{NYT10m}
The original NYT10~\cite{riedel2010modeling} dataset, constructed using Freebase~\cite{freebase} in a distantly-supervised manner, allows for locating all entities in NYT10 within Freebase by their identifiers. However, Freebase was discontinued in 2014, preventing the extraction of specific knowledge contexts. Consequently, we utilized FB15k~\cite{bordes2013translating}, a Freebase~\cite{freebase} subset, as a knowledge base to provide knowledge context for NYT10m. Specifically, we linked all entities in NTY10m to their counterparts in FB15k based on identifiers and extracted relation facts containing these linked counterparts (entities) to construct our KG.

\xhdr{Wiki80 and Wiki20m}
The original Wiki80 and Wiki20~\cite{han-etal-2020-data} datasets were created by aligning entities in the English Wikipedia corpus with Wikidata~\cite{wikidata} which provides Wiki identifiers (unique identifiers for entities in Wikidata) for all entities in the two datasets. As a result, we directly pinpointed entities in Wikidata using the entities' Wiki identifiers~\footnote{Wikidata is a real-time updated knowledge graph, and we obtained our data on 7 Mar 2022.}. Then, we extracted relational facts containing the entities in Wikidata to construct auxiliary KGs for Wiki80 and Wiki20m.

\subsection{Dataset Refinement}
\label{sec:refinement}
As previous datasets suffer from data quality issues and lack unified auxiliary KGs, dataset refinement aims to enhance the data quality of each RE dataset and construct KGs for datasets in accordance with the knowledge base.

To provide high-quality datasets, we denoised all the original datasets. We updated identifiers of redirected entities to the latest version and removed instances containing missing entities in Wiki80~\cite{han-etal-2019-opennre} and Wiki20m~\cite{gao-etal-2021-manual}. Subsequently, we eliminated duplicate instances in each dataset. We discovered that some relation facts in Wiki80 and Wiki20m had the subject and object referring to the same entity, such as:
\begin{equation*}
(soviets [Q15180],sovietunion [Q15180],~country [P17]),
\end{equation*}
and we removed instances containing these relational facts. Eventually, in Wiki80, there were 162 redirected identifiers, and 32 entities were removed; in Wiki20m, there were 1,092 redirected identifiers, and 269 entities were removed.

\xhdr{KG for NYT10m}
NYT10m~\cite{gao-etal-2021-manual} only provides instances with relational facts, but no KG context for their entities. Thus, we constructed an informative KG for this dataset, enabling it to serve as a benchmark for KG-enhanced RE tasks. After linking all the entities to FB15k as elaborated in Section~\ref{sec:linking}, we extracted triplets of these entities from the knowledge base to construct the KG.

Formally, let $\mathcal{E}$ be a set of entities containing all entities in NYT10m and let $\mathcal{R}$ be a set of relations containing all relations in NYT10m. Given every triplet $(h^j, t^j, r^j)$ in FB15k, where $h^j$ and $t^j$ are subject and object entities with relation $r^j$, we construct the KG using every triplet that satisfies the following rules: (1) $h^j$ and $t^j$ can be found in $\mathcal{E}$, (2) $r^j$ is a relation in $\mathcal{R}$. In addition, for every instance $r^i=(x^i, e^i_1, e^i_2)$ in NYT10m, where $e^i_1$ and $e^i_2$ delimit entity mentions in token sequence $x^i$, we can extract a triplet such as $(e^i_1, e^i_2, r^i)$. These triplets can be added to the KG to avoid information shortage. In the final KG, the triplets extracted from validation/test sets must be excluded.

\xhdr{KGs for Wiki80 and Wiki20m}
Similarly, both Wiki80 and Wiki20m only provide instances with relational facts. We also extract suitable information from Wikidata~\cite{wikidata} for KG construction. The KG construction method we employed for Wiki80 and Wiki20m is the same as the one we used for NYT10m.

\section{Data Analysis}\label{data analysis}
In this section, we provide the statistics of KERED and examine the extent to which these datasets facilitate relation extraction tasks augmented by knowledge graphs.

\subsection{Analysis of Instances}
As discussed in Section~\ref{sec:refinement}, dataset refinement eliminates noise from the original DSRE datasets to ensure data quality. 
Consequently, we obtained a modified version of NYT10m with 475,401 instances, Wiki80 with 55,547 instances, and Wiki20m with 743,703 instances. Wiki80 is smaller than the other datasets, while the scale of NYT10m is comparable to that of Wiki20m. 
However, the number of entities and relational facts in Wiki20m is considerably higher than those in NYT10m, and the N/A proportion of NYT10m is higher than that in Wiki20m. 
The statistics of our modified datasets in KERED are presented in Table~\ref{tab:statistics_instance}.

\begin{table*}[ht]
\small
\centering

\begin{threeparttable}[b]
\begin{tabular}{ccccccccc}
\toprule
\multicolumn{2}{c}{Dataset}      & Manual  & Instances & Entities & Facts & N/A & Relations           & KB                        \\ \midrule
\multirow{3}{*}{NYT10m}  & train & No      &417,893     &61,112     &17,137     &80\%     & \multirow{3}{*}{25} & \multirow{3}{*}{Freebase} \\
                         & valid & No      &46,422      &20,850     &4,062      &80\%     &                     &                           \\
                         & test  & Part     &11,086      &4,554      &3,899      &28\%     &                     &                           \\ \midrule
\multirow{2}{*}{Wiki80} & train & All  &50,353  &66,758  &50,353  &0\% & \multirow{2}{*}{80} & \multirow{2}{*}{Wikidata} \\
                         & val  & All     &5,194   &8,662  &5,194  &0\%     &     &       \\ \midrule
\multirow{3}{*}{Wiki20m} & train & No      &571,787  &285,905  &154,078  &55\%  & \multirow{3}{*}{81} & \multirow{3}{*}{Wikidata} \\
                         & valid & No      &48,794   &44,082   &16,489   &66\%  &                     &                           \\
                         & test  & All    &123,122  &89,925   &53,755   &23\%  &                     &                           \\
\bottomrule
\end{tabular}



\caption{Statistics of KERED datasets.
The \emph{Manual} column indicates whether the instances are manually labeled.
\emph{Instances}, \emph{Entities}, and \emph{Facts} indicate the number of instances, entities, and relational facts, respectively.
The \emph{N/A} column shows the percentages of instances with N/A relation in the datasets.
\emph{Relation} indicates the number of relations in each dataset.
KB indicates the knowledge base source of datasets in KERED.
Wiki80 and Wiki20m are our modified versions.}
\label{tab:statistics_instance}

\end{threeparttable}
\end{table*}

\subsection{Analysis of Knowledge Graphs}~\label{sec:akg}
In Section~\ref{sec:refinement}, we detailed the process of KG generation for each DSRE dataset. For NYT10m, we used a subset of FB15k \cite{bordes2013translating} as the KG, which covers approximately 25\% of entities in NYT10m. For Wiki80 and Wiki20m, we extracted the knowledge context of entities from Wikidata \cite{wikidata} and generated KGs with this context. These KGs contain a large number of relational facts and cover almost all entities in Wiki80 and Wiki20m.

The aforementioned KGs provide abundant auxiliary information for KG-enhanced relation extraction tasks. To enhance their performance, knowledge-enhanced RE methods can train knowledge embeddings with our KGs or consider relational facts in the KGs as input \cite{recon, xu-barbosa-2019-connecting, lin2022multimodal}.

We present the statistics of KGs in KERED in Table~\ref{tab:statistics_kg}, wherein the numbers of entities and facts reflect the scales of KGs. On one hand, the scales of the Wiki20m dataset and KG are relatively large, indicating that Wiki20m is suitable for evaluating the performance of knowledge-enhanced RE methods. On the other hand, the scales of the Wiki80 dataset and KG are small, suggesting that we can quickly assess the performance of models with Wiki80. Moreover, the scale of the KG for NYT10m is considerably smaller than that of the NYT10m dataset, and this KG can only provide knowledge context for approximately 25\% of the entities in NYT10m. The degree of an entity is the number of triples containing that entity in the KG. The average degree of all entities reflects the density of the KG. For Wiki80 and Wiki20m, the KGs provide 7.98 and 8.60 direct neighbors for each entity on average, indicating that the information in these KGs is dense. Because KG embeddings are sensitive to the density of the KG \cite{pujara-etal-2017-sparsity}, we believe that our dense KGs are beneficial to KG-enhanced RE methods, which often rely on KG embeddings \cite{recon, xu-barbosa-2019-connecting, lin2022multimodal, sun-etal-2020-colake}. For NYT10m, the average degree is 2.87, which is relatively low but acceptable. According to the number of connected components and the size of the maximal connected components, each KG consists of a very large connected component and a few small connected components. Relational information in large connected components is denser than the information in small ones. As the maximal component contains almost all entities in each KG, KG-enhanced methods can effectively extract features of entities and relations from these KGs.

\begin{table*}[ht]
\small
\centering

\begin{threeparttable}[b]

\begin{tabular}{ccccccccc}
\toprule
Dataset & Total En. & KG En. & Facts & Degree & Comp. & Max Comp. & KB \\ \midrule
NYT10m  & 64,890 & 16,469 & 23,643 & 2.87 & 733 & 14,730  & FB15k \\
Wiki80  & 72,358 & 72,353 & 288,750 & 7.98 & 513 & 70,477 & Wikidata \\
Wiki20m & 360,966 & 360,956 & 1,551,694 & 8.60 & 621 & 359,145 & Wikidata \\
\bottomrule
\end{tabular}

\caption{Statistics of KG for each dataset in KERED. \textit{Total En.} is the total number of entities in the dataset. \textit{KG En.} and \textit{Facts} show the number of entities and relational facts in the KG (all entities and relations in the KG are also in the dataset). \textit{Degree} is the average degree of all entities in the KG. \textit{Comp.} is the number of connected components in the graph. \textit{Max Comp.} is the size of the maximal connected component. \textit{KB} is the knowledge base of each KG.}
\label{tab:statistics_kg}

\end{threeparttable}
\end{table*} 

\begin{figure*}[ht]
    \centering
    \includegraphics[width=\textwidth]{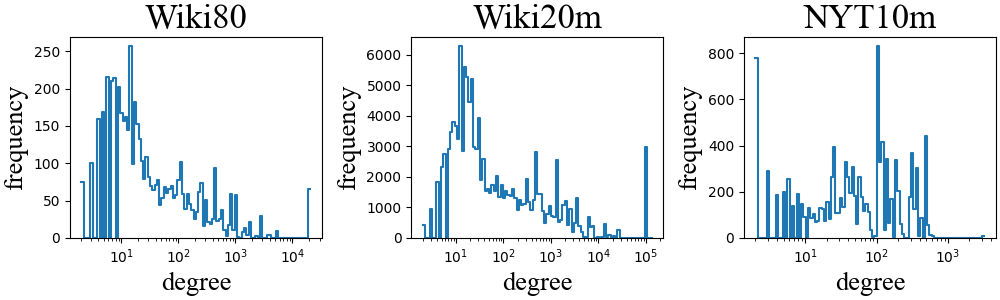}
    \caption{Frequency distributions for degrees of instances.
    The X-axis represents the number of degrees, and the Y-axis represents the log-scaled frequency counts of instances.
    The degree of each instance is obtained from the sum of degrees of subject and object entities in KG.
    }
    \label{fig:degrees}
\end{figure*}

We further analyze the statistics of the KGs from the perspective of instances.
As shown in Figure \ref{fig:degrees}, for Wiki80 and Wiki20m, the distributions are generally similar, and the frequency of instances reaches its peak when the degree is approximately 16, after which the frequency tends to decrease with the increase in the degree.
For NYT10m, the frequency of instances is high when the degree is 2, indicating that many entities only have one neighbor entity in the KG.
In general, the distribution is similar between Wiki80 and Wiki20m.
The difference is that the amount of data in Wiki80 is small, so the overall log-scaled frequency is small.
However, the distribution of NYT10m is quite different; it is sparse and irregular.

\section{Experimental Settings}\label{benchmarks}
To assess the performance of existing RE methods, particularly knowledge-enhanced RE methods, we carry out comprehensive experiments. 
In this section, we describe our benchmarks and experimental settings.

\subsection{Benchmarks}
We assembled various current RE methods as our benchmarks to assess their performance on our datasets. 
Below, we offer a brief overview of all RE benchmarks that we evaluated on our KG-enhanced datasets, including RE methods without KG, sentence-level methods with KG, and bag-level RE methods with KG.

All methods underwent supervised training or fine-tuning on the training set and were evaluated on the validation/test set.
For PCNN \cite{zeng-etal-2015-distant} and BERT \cite{devlin-etal-2019-bert}, we adhered to the experimental settings of \cite{gao-etal-2021-manual, han-etal-2019-opennre}.
For the other benchmarks, we employed their original settings and hyperparameters, except for the training epochs.

\subsubsection{RE Methods without KG}
We chose two widely-used RE models, PCNN \cite{zeng-etal-2015-distant} and BERT \cite{devlin-etal-2019-bert}, as our benchmarks.
These methods classify relations between entities without utilizing KG information.
We conducted our experiments on the original Wiki80, Wiki20m, and NYT10m or referenced the experimental results from \cite{gao-etal-2021-manual}.

\begin{itemize}[leftmargin=1em]
\setlength\itemsep{0em}
\item \textbf{PCNN} \cite{zeng-etal-2015-distant} is a DSRE method developed to automatically learn features from evidence sentences.
It employs piece-wise convolutional neural networks to encode words in a sentence and generate sentence-level embeddings.

\item \textbf{BERT} \cite{devlin-etal-2019-bert} and its variants have emerged as a promising natural language processing approach since 2019.
This pretraining method achieves remarkable performance in various tasks including relation extraction.

\end{itemize}

\subsubsection{Sentence-level RE Methods with KG}
Sentence-level RE methods leverage the auxiliary KG for RE tasks in various ways.
The most common approach is the use of embeddings derived from KGs.
We selected three methods to evaluate their sentence-level performance on our proposed dataset.

\begin{itemize}[leftmargin=1em]
\setlength\itemsep{0em}
\item \textbf{RECON} \cite{recon} employs a graph neural network to learn representations of both relational facts in KG and the evidence sentences.
Attributes such as entity names, aliases, and types from KG entities are also utilized.

\item \textbf{KGPool} \cite{nadgeri-etal-2021-kgpool} dynamically selects KG context with graph pooling to enhance RE performance.
This method considers only entity attributes as side information from the KG, rather than relational facts.

\item \textbf{CoLAKE} \cite{sun-etal-2020-colake} utilizes an unlabeled data structure called word-knowledge graph, integrating language and knowledge context to learn contextualized representations.
In the fine-tuning process, we used the label \emph{unk} to represent the N/A relation in Wiki20m.
\end{itemize}

\subsubsection{Bag-level RE Methods with KG}
Bag-level RE methods classify relations between entities at the bag level rather than the sentence level.
We evaluated three bag-level methods on KERED.

\begin{itemize}[leftmargin=1em]
\setlength\itemsep{0em}
\item \textbf{CGRE} \cite{liang2021distantly} constructs a constraint graph with entity type information to reflect dependencies between relations.
It generates representations for entities and relations by encoding the graph into vectors and extracting features of the nodes to support RE tasks.

\item \textbf{HRERE} \cite{xu-barbosa-2019-connecting} is a neural framework that jointly learns knowledge and language representations.
It integrates relation extraction and KG embedding generation in a unified manner to enhance performance.

\item \textbf{REMAP} \cite{lin2022multimodal} is a multimodal method for the DSRE task.
It fuses knowledge graph embeddings with deep language models to effectively extract and classify the relations between entities.
\end{itemize}

\subsection{Metrics}
We use micro F1 and micro average precision (AP) as metrics in our experiment.
Micro F1 is the harmonic mean of global micro precision and global micro recall.
The formulas for micro F1 are:
\begin{equation}
    \textrm{F1} = \frac{2*\textrm{precision}*\textrm{recall}}{\textrm{precision}+\textrm{recall}},
\end{equation}
\begin{equation}
    \textrm{precision} = \frac{\textrm{TP}}{\textrm{TP} + \textrm{FP}},
\end{equation}
\begin{equation}
    \textrm{recall} = \frac{\textrm{TP}}{\textrm{TP} + \textrm{FN}},
\end{equation}
where $TP$ is the global true positive rate, $FP$ is the global false positive rate, and $FN$ is the global false negative rate.
AP is the weighted mean of precision at each threshold for all samples.
The formula for micro AP is:
\begin{equation}
    \textrm{AP} = \sum_{i=2}^{n} \textrm{precision}_{i}*(\textrm{recall}_{i} - \textrm{recall}_{i-1}),
\end{equation}
where $\textrm{precision}{i}$ and $\textrm{recall}{i}$ represent the global precision and recall at the $i$th threshold, and $n$ denotes the total number of samples.
The Average Precision (AP) is also referred to as the Area Under the Curve (AUC) of the Precision-Recall (PR) curve in some articles \cite{gao-etal-2021-manual, han-etal-2019-opennre}.

In the case of sentence-level RE, the relation for each evidence sentence is predicted directly, and the results are reported on the test sets.
In the case of bag-level RE, instances sharing the same entity pairs were grouped into bags, and a prediction is made for each bag in the test sets.
Following the configuration of most RE experiments \cite{nadgeri-etal-2021-kgpool, gao-etal-2021-manual, recon, han-etal-2019-opennre, xu-barbosa-2019-connecting}, all instances, including those with N/A labels, are considered for micro F1 calculation.
Conversely, instances with N/A labels are excluded from the micro AP calculation.

\section{Results and Discussion}\label{discussion}
In the previous section, the experimental settings for our benchmarks were delineated.
This section presents the results of our experiments, adhering to the established configurations.
Additionally, an analysis of the results is conducted, along with a discussion on the influence of KG information on RE tasks.

\subsection{Experimental Results}
\label{sec:results}

\begin{table*}[ht]
\centering
\setlength{\tabcolsep}{3.4pt}

\begin{threeparttable}[b]

    {\fontsize{9}{10}\selectfont
    
    \begin{tabular}{p{1.5mm}cccccccc}
    \toprule
    &\multirow{2}{*}{Model}& \multicolumn{2}{c}{Wiki80} & \multicolumn{2}{c}{Wiki20m} & \multicolumn{2}{c}{NYT10m} \\
     && F1 & AP & F1 & AP & F1 & AP  \\
    \midrule
    \cellcolor{blue!10} &PCNN+AVG\tnote{$\dagger$}  &--- &--- &\emph{71.8} &\emph{78.1} &\emph{53.6} &\emph{52.9}  \\
    \cellcolor{blue!10} &PCNN+ONE\tnote{$\dagger$}  &77.43 &84.68 &\emph{70.3} &\emph{76.6} &\emph{54.8} &\emph{53.4}  \\
    \cellcolor{blue!10} &PCNN+ATT\tnote{$\dagger$}  &--- &--- &\emph{71.2} &\emph{77.5} &\emph{56.5} &\emph{56.8}  \\
    \cellcolor{blue!10} &BERT+AVG\tnote{$\dagger$}  &--- &--- &\textbf{82.7} &\textbf{89.9} &\emph{60.4} &\emph{56.7}  \\
    \cellcolor{blue!10} &BERT+ONE\tnote{$\dagger$} &\textbf{86.69} &\textbf{93.41} &\emph{81.6} &\emph{88.9} &\textbf{61.9} &\textbf{58.1}  \\
    \cellcolor{blue!10}\multirow{-6}{*}{\rotatebox[origin=c]{90}{\textit{RE w.o. KG}}}& BERT+ATT\tnote{$\dagger$}  &--- &--- &\emph{66.8} &\emph{70.9} &\emph{54.1} &\emph{51.2}  \\
    \midrule
    \cellcolor{blue!10} &RECON~\cite{recon}  & 77.58$\pm$0.30 &86.86$\pm$0.17 &74.85$\pm$1.02 & 92.58$\pm$0.68 & \textbf{52.09$\pm$1.58} & \textbf{68.30$\pm$2.30}  \\
    \cellcolor{blue!10} &KGPool~\cite{nadgeri-etal-2021-kgpool} & 78.13$\pm$0.39 & 87.51$\pm$0.23 & 78.61$\pm$0.25 & 89.67$\pm$0.23 & 50.78$\pm$0.83 & 64.18$\pm$0.18 \\
    \cellcolor{blue!10}\multirow{-3}{*}{\rotatebox[origin=c]{90}{\textit{S. RE}}} &CoLAKE~\cite{sun-etal-2020-colake}\tnote{$\ddag$}  & \textbf{91.78$\pm$0.08} & \textbf{96.89$\pm$0.03} & \textbf{86.05$\pm$0.77} & \textbf{92.74$\pm$0.79} & --- & ---\\
    \midrule
    \cellcolor{blue!10} &CGRE~\cite{liang2021distantly} &--- &--- &--- &--- & 51.48$\pm$0.66 & 53.51$\pm$0.40 \\
    \cellcolor{blue!10} &HRERE~\cite{xu-barbosa-2019-connecting}  & 80.04$\pm$0.21 & 87.42$\pm$0.17 & 77.33$\pm$0.29 & 88.38$\pm$0.24 & 30.50$\pm$0.16 & 54.40$\pm$0.88 \\
    \cellcolor{blue!10}\multirow{-3}{*}{\rotatebox[origin=c]{90}{\textit{B. RE}}}  &REMAP~\cite{lin2022multimodal}  & \textbf{88.73$\pm$0.18} & \textbf{92.16$\pm$0.22} & \textbf{84.99$\pm$0.21} & \textbf{89.92$\pm$0.19} & \textbf{52.19$\pm$0.92} & \textbf{66.39$\pm$1.19}\\
    \bottomrule
    \end{tabular}

    \begin{tablenotes}
        \item[$\dagger$] These experiments were conducted on original RE datasets under the settings of \cite{gao-etal-2021-manual, han-etal-2019-opennre}. We conducted the experiments for Wiki80, and the results of these models on Wiki20m and NYT10m have been cited from \cite{gao-etal-2021-manual}.
        
        \item[$\ddag$] Because CoLake is pre-trained on Wikidata, there is a risk of information leakage.\\~
    \end{tablenotes}
    }
    
    \caption{Experimental results (\%) of baselines. We report three-run micro F1 scores and micro average precision (AP) scores with standard deviations. \emph{RE w.o. KG} indicates RE methods without using KG. \emph{S. RE} and \emph{B. RE} indicate sentence-level and bag-level RE methods with KG. For PCNN and BERT, we only evaluate PCNN+ONE and BERT+ONE because the other two aggregation methods do not fit the sentence-level dataset Wiki80. CoLAKE was left untested on NYT10m because of the lack of learned relation representations. CGRE was only evaluated on NYT10m because the constraint graph it relies on only contains relations from Freebase.}
    \label{tab:Benchmarks}

\end{threeparttable}

\end{table*}

Table \ref{tab:Benchmarks} shows our experimental results.
RE methods without using KG are evaluated on original Wiki80, Wiki20m, and NYT10m datasets; Sentence-level and bag-level methods with KG are evaluated on KERED.

Average (AVG), at-least-one (ONE), and attention (ATT) represent methods employed in bag-level Relation Extraction (RE) for aggregating sentence-level prediction results into bag predictions \cite{gao-etal-2021-manual}.
AVG calculates the mean of all sentence representations within the bag.
ONE predicts relation scores for each sentence in the bag and subsequently selects the highest score for each relation.
ATT computes a weighted average over sentence embeddings in the bag, determining weights through attention scores between sentences and relations.
Because Wiki80 is a sentence-level RE dataset without any repetitive entity pairs, AVG and ATT are inapplicable to experiments on this dataset.

Because CoLAKE is a pre-trained model based on Wikidata5M \cite{wang-etal-2021-kepler-custom}, it lacks learned relation representations in NYT10m.
Therefore, we only evaluate the method on Wiki80 and Wiki20m.
Additionally, some triplets in the test set may appear in the training data of CoLAKE, causing information leakage.
The method CGRE is only evaluated on NYT10m because the constraint graph provided by the author only contains relations from Freebase \cite{freebase} rather than Wikidata \cite{wikidata}.

The experimental results in Table \ref{tab:Benchmarks} lead to the following observations:
Among RE methods without KG information, BERT+AVG exhibits the best performance on Wiki20m, attaining 82.7\% for F1 and 89.9\% for AP.
BERT+ONE demonstrates superior performance on NYT10m and Wiki80.
For sentence-level RE methods incorporating KG, CoLAKE significantly outperforms other models on Wiki80 and Wiki20m, achieving 91.78\% for F1 and 96.89\% for AP on Wiki80, and 86.05\% for F1 and 92.74\% for AP on Wiki20m.
RECON is the top sentence-level performer on NYT10m, reaching 52.09\% for F1 and 68.30\% for AP.
Among bag-level RE methods with KG information, REMAP attains the best performance for both metrics across all three datasets.

Our experimental results indicate that RE methods incorporating KG generally surpass those without KG, particularly in AP scores.
The high performance can be attributed to the additional KG information supplying more knowledge to RE models.
A notable exception is the poor performance of all RE methods with KG in micro F1 scores on NYT10m.
This could be due to the limited number of relational facts in the KG constructed for NYT10m.
Furthermore, other KG information such as entity attributes employed by RECON and KGPool, or constraint graphs utilized by CGRE, may not offer the models a beneficial predictive direction.
Another observation is that methods leveraging pre-training exhibit better performance compared to other methods.

\subsection{Analysis and discussion}
We conduct a comprehensive analysis of the experiments and then discuss the effect of our curated KG, including basic statistics from the perspective of instances and the effect of KG on model performance.

\subsubsection{Analysis of Performance and Degree of Instance}
\label{sec:degree_and_instance}
\begin{figure*}[ht]
    \centering
    \includegraphics[width=\textwidth]{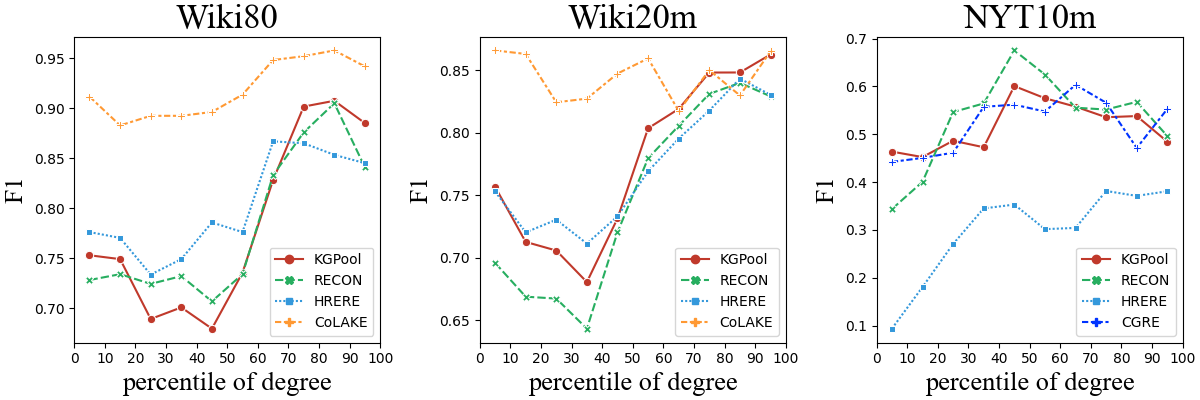}
    \caption{Micro F1 scores of RE models on instances with different percentiles of degree.
    The degree of each instance is obtained from the sum of the degrees of the subject and object entities in KG.
    We split the instances into ten groups according to their degrees.
    Degrees are maximum when the percentile of degree is 95 and minimum when it is 5.}
    \label{fig:pod}
\end{figure*}

In Section~\ref{sec:akg}, we analyzed the statistics of KG and presented the frequency distributions for degrees of instances in Table~\ref{fig:degrees}.
Herein, We further analyze the influence of instances' degree on the performance of models.
As shown in Figure \ref{fig:pod}, with the increment of degree, micro F1 scores of KGPool, RECON, and HRERE initially decline and subsequently rise in Wiki80 and Wiki20m.
In contrast, this trend is not evident in NYT10m, attributable to its limited built KG.
As highlighted in Section \ref{sec:results}, the pre-training method CoLAKE outperforms other approaches, especially when the percentile degree is lower than 70\%.
Furthermore, the performance of CoLAKE and CGRE are not sensitive to the number of degrees, as the methods do not utilize relational facts in our curated KGs.

\subsubsection{Analysis of Performance and Relation}
To investigate model performance on individual relations, we present micro F1 scores of models for each relation in Figure \ref{fig:rel} (see Appendix B).
The results demonstrate considerable variation in the ability of models to accurately classify each relation.
Some relations are readily identifiable, such as $voice type$ and $position played on team/specialty$ in Wiki80, while others are not.
Additionally, micro F1 scores can reach $1.0$ for certain relations in Wiki80 and Wiki20m, but this observation does not extend to the NYT10m dataset.
Meanwhile, some relations in NYT10m remain unrecognizable by the models, which may be attributed to the limited training data available for these relations.

\subsubsection{Discussion}
Based on the analyses presented, it can be concluded that the availability of relational facts plays a crucial role in knowledge-enhanced RE tasks.
Due to the relatively small KG of NYT10m, the performance of most methods with KG is inferior to those without KG, indicating that these models may rely excessively on KG information in their design, potentially causing a negative impact when KG information is insufficient.
Predicting relations between entities can be challenging, particularly when the quantity of relational facts in KG is limited.
Thus, although knowledge-enhanced RE methods are effective, we believe that knowledge-enhanced RE merits further investigation.
We propose the following research directions: (1) developing models that effectively utilize KG information, particularly relational facts; and (2) investigating approaches that capitalize on limited relational facts to enhance RE performance.

\section{Conclusion}\label{conclusion}
We developed KERED using three knowledge-enhanced RE datasets derived from widely-used DSRE datasets.
We identified issues within the original datasets, subsequently cleaned them to enhance data quality, and linked entities to large-scale knowledge bases.
Our curated KGs in KERED promote the study of knowledge-enhanced RE methods in both sentence-level and bag-level settings.
Furthermore, we established a connection between auxiliary KGs and knowledge-enhanced RE methods, demonstrating that our auxiliary KGs consistently benefit these approaches.
In conclusion, we provided three high-quality datasets for fair comparisons between knowledge-enhanced RE methods and established benchmarks in various settings, which will further advance RE research.

\section*{Acknowledgements}
This work was supported by National Natural Science Foundation of China (No.72001205, 72171229, 11971504), College Stability Support Program of Shenzhen (20200827090247001), ~fund for building world-class universities (disciplines) of Renmin University of China, Chinese National Statistical Science Research Project (2022LD06), Foundation from Ministry of Education of China (20JZD023), Ministry of Education Focus on Humanities, and Social Science Research Base (Major Research Plan 17JJD910001).

\section*{Compliance with Ethical Standards}
\textbf{Conflict of interest} The authors declared that they have no conflicts of
interest with regard to this work. We declare that we do not have any commercial or associative interest that represents a conflict of interest in connection with the work submitted.

\begin{appendices}

\section{Access to KERED} \label{sec:access}
We have posted our refined datasets Wiki80, Wiki20m, and NYT10m on Figshare: \url{https://figshare.com/projects/KERED/134459}.
Each dataset consists of four major components: a knowledge graph, training set, validation set, and a testing set (since the original Wiki80 has no testing set, we consider the validation set as the testing set for our experiments).
The knowledge graphs are in CSV format.
Each entry in the knowledge graphs represent a relational fact (a subject, an object, and a relation).
For training/validation/testing sets, we follow the format of the original datasets.
All instances in these sets are in json format, providing sentences, relations, and entities by key/value pairs.

\newpage
\setcounter{figure}{0}
\renewcommand{\thetable}{B.\arabic{table}}
\renewcommand{\thefigure}{B.\arabic{figure}}
\section{Model Performance on Relations}
\label{sec:top-10}

\begin{figure*}[!h]
    \centering
    \subfigure[HRERE]{
    \includegraphics[width=0.78\linewidth]{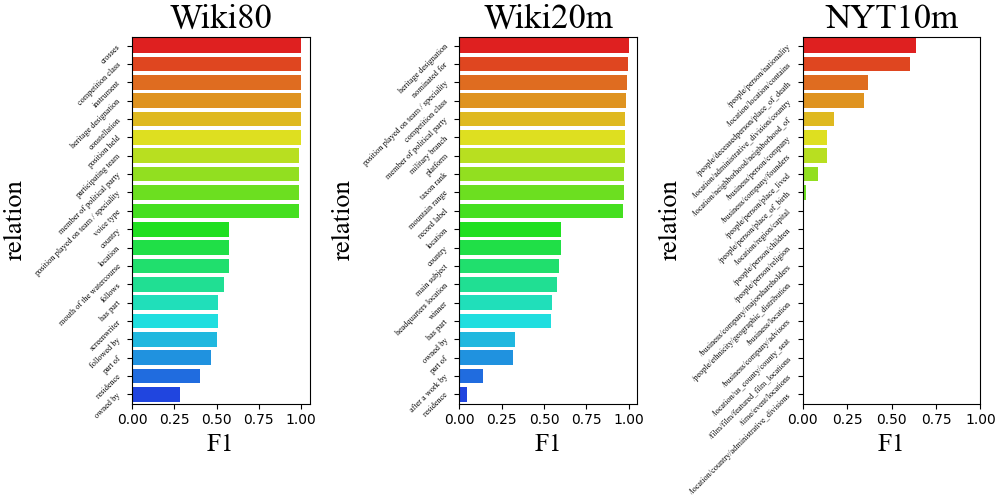}
    \label{fig:rel_hrere}}
    \subfigure[KGPool]{
    \includegraphics[width=0.78\linewidth]{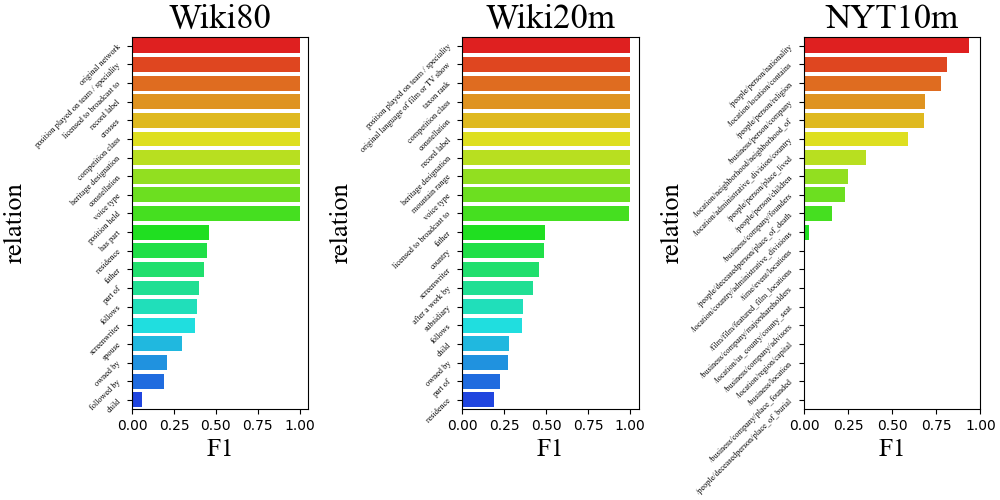}
    \label{fig:rel_kgpool}}
    \subfigure[RECON]{
    \includegraphics[width=0.78\linewidth]{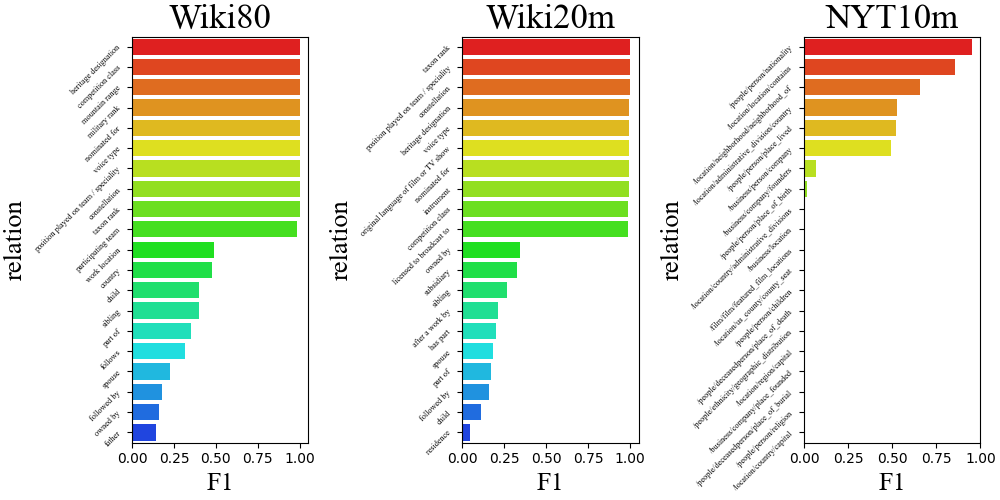}
    \label{fig:rel_recon}}
    \caption{Top-10 high and top-10 low micro F1 scores of different relations for HRERE, KGPool, and RECON.}
    \label{fig:rel}
\end{figure*}

\end{appendices}


\bibliography{anthology,custom}

\end{document}